\documentclass[runningheads]{llncs}
\usepackage{graphicx}
\usepackage{amsmath}
\usepackage{multirow}
\usepackage{amssymb}
\usepackage{cite}
\usepackage{hyperref}
\def\bs{\boldsymbol}
\begin{document}
\title{Defense-VAE: A Fast and Accurate Defense against Adversarial Attacks}
%\author{Xiang Li\inst{1}\orcidID{0000-1111-2222-3333} \and
%Shihao Ji\inst{1}\orcidID{1111-2222-3333-4444} }
\author{Xiang Li, \and Shihao Ji}
\authorrunning{X. Li, S. Ji}
\institute{
    Georgia State  University, USA \\
    \email{xli62@student.gsu.edu, sji@gsu.edu}}
\maketitle              % typeset the header of the contribution
\begin{abstract}
Deep neural networks (DNNs) have been enormously successful across a variety of prediction tasks. However, recent research shows that DNNs are particularly vulnerable to adversarial attacks, which poses a serious threat to their applications in security-sensitive systems. In this paper, we propose a simple yet effective defense algorithm Defense-VAE that uses variational autoencoder (VAE) to purge adversarial perturbations from contaminated images. The proposed method is generic and can defend white-box and black-box attacks without the need of retraining the original CNN classifiers, and can further strengthen the defense by retraining CNN or end-to-end finetuning the whole pipeline. In addition, the proposed method is very efficient compared to the optimization-based alternatives, such as Defense-GAN, since no iterative optimization is needed for online prediction. Extensive experiments on MNIST, Fashion-MNIST, CelebA and CIFAR-10 demonstrate the superior defense accuracy of Defense-VAE compared to Defense-GAN, while being 50x  faster than the latter. This makes Defense-VAE widely deployable in real-time security-sensitive systems. Our source code can be found at \url{https://github.com/lxuniverse/defense-vae}.
\end{abstract}

\section{Introduction}
Deep neural networks (DNNs) have demonstrated remarkable success in solving complex prediction tasks. However, recent studies show that they are particularly vulnerable to adversarial attacks~\cite{biggio2013evasion, papernot2016transferability, szegedy2013intriguing} in the form of small perturbations to inputs that lead DNNs to predict incorrect outputs. For images, such perturbations are often almost imperceptible to human vision system, while being very effective at fooling DNN-based systems. Both white-box attacks \cite{Papernot_2016} and black-box attacks~\cite{Papernot_2017} have been proposed to attack DNNs, and they can often fool the network with high probabilities. These attacks pose a serious threat to the applications of DNNs in security-sensitive systems, e.g., identity authentication surveillance, self-driving cars, malware detection, and voice command recognition. As a result, it is critical to develop effective and efficient defense mechanisms to counter adversarial attacks.

\begin{figure}[t]	
 	\begin{center}
 		\includegraphics[width=0.95\linewidth]{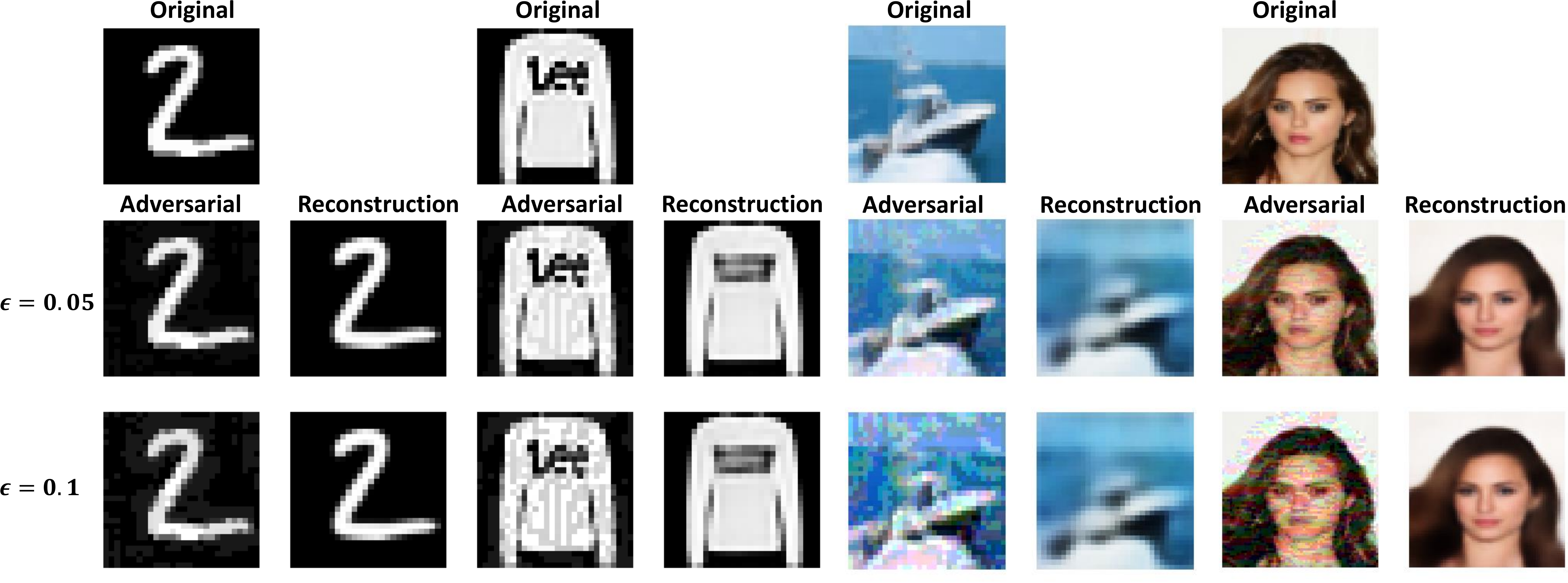}
 	\end{center}\vspace{-8pt}
 	\caption{Defense-VAE purges adversarial perturbations from contaminated images. Example images are from MNIST, F-MNIST, CIFAR-10, and  CelebA, respectively. FGSM~\cite{goodfellow6572explaining} with $\epsilon=0.05$ and $\epsilon=0.1$ are used to generate the adversarial attacks.}
 	\label{fig:intro}
 \end{figure}

In this paper, we propose a simple yet effective defense mechanism called Defense-VAE that uses Variational AutoEncoder (VAE)~\cite{kingma2013auto,vae_rezende} to purge the adversarial perturbations from contaminated images before feeding the images to the downstream CNN classifiers. To illustrate the idea, we generate some adversarial images based on the FGSM attack~\cite{goodfellow6572explaining} with $\epsilon=0.05$ and $\epsilon=0.1$ on four popular image classification benchmarks: MNIST~\cite{mnist}, Fashion-MNIST~\cite{xiao2017fashion}, CIFAR-10~\cite{cifar10} and CelebA~\cite{liu2015faceattributes}. These adversarial images are then fed into Defense-VAE for reconstruction. Figure~\ref{fig:intro} illustrates some of the typical examples from Defense-VAE. As we can see, the Defense-VAE generated images are the faithful reconstructions from the underlying clean images, with the majority of adversarial perturbations removed. As we will demonstrate later, such reconstructed images can recover almost all the accuracy losses due to adversarial attacks, without introducing much computation overhead compared to Defense-GAN~\cite{samangouei2018defensegan}, a closely related state-of-the-art defense algorithm that is based on Generative Adversarial Networks (GAN)~\cite{gan}. 

Compared with the state-of-the-art defense algorithms, our method has the following properties:
\begin{itemize}
	\item Defense-VAE is very generic and can defend white-box attacks and black-box attacks without the need of retraining the original CNN classifiers, and can further strengthen the defense by retraining or end-to-end finetuning;
	\item Defense-VAE achieves much higher accuracy than the state-of-the-art defense algorithms on white-box and black-box attacks. Especially, it outperforms Defense-GAN by about 30\% in defending black-box attacks on F-MNIST;
	\item Defense-VAE is very efficient compared to the optimization-based alternatives, such as Defense-GAN, as no iterative optimization is needed for online prediction. From our experiments, it shows that Defense-VAE is about 50x faster than Defense-GAN. This makes our method widely deployable in real-time security-sensitive applications.
\end{itemize}

%------------------------------------------------------------------------
\section{Defense-VAE: The Proposed Algorithm}
At a high level, Defense-VAE is a defense algorithm that is based on deep generative models for image reconstruction. That is, given an adversarial image as input, the generative model attempts to produce a denoised image that is closely related to the underlying clean image, with the adversarial perturbations removed. As the name suggested, Defense-VAE is built upon Variational AutoEncoder (VAE)~\cite{kingma2013auto,vae_rezende}. Therefore, we first give a brief introduction to VAE.

\subsection{Variational Auto-Encoder}

Variational Autoencoder (VAE)~\cite{kingma2013auto,vae_rezende} is one of the most powerful deep generative models that is based on latent variable models. It consists of an encoder network to encode an input image to the latent variable $\bs{z}$ and a decoder network to decode the latent variable $\bs{z}$ back to the image domain:
\begin{align}
\boldsymbol{z} \sim \operatorname { Enc } ( \boldsymbol { x } ) = q ( \boldsymbol { z } | \boldsymbol { x } ) , \quad \boldsymbol { x } \sim \operatorname { Dec } ( \boldsymbol { z } ) = p ( \boldsymbol { x } | \boldsymbol { z } ).
\end{align}
Since the maximum likelihood (ML) estimate of this latent variable model is intractable, a variational lower bound (ELBO) is optimized instead:
\begin{align}
\mathcal { L } _ { \mathrm { VAE } } &= - \mathbb { E } _ { q ( \boldsymbol { z } | \boldsymbol { x } ) } \left[ \log \frac { p ( \boldsymbol { x } | \boldsymbol { z } ) p ( \boldsymbol { z } ) } { q ( \boldsymbol { z } | \boldsymbol { x } ) } \right]\\ \nonumber
& = - \mathbb { E } _ { q ( \boldsymbol { z } | \boldsymbol { x } ) } [ \log p ( \boldsymbol { x } | \boldsymbol { z } ) ]  +
D _ { \mathrm { KL } } ( q ( \boldsymbol { z } | \boldsymbol { x } ) \| p ( \boldsymbol { z } ) )
\end{align}
where the first term is the reconstruction error and the second term is a regularization that prefers the posterior to be close to the prior. Typically, a simple unit Gaussian prior is assumed in VAE. To facilitate efficient computation, a diagonal covariance Gaussian posterior is further assumed, which enables the use of the reparameterization trick to reduce the variance of Monte-Carlo sampling~\cite{kingma2013auto}. 

As a generative model, VAE can generate high quality images that follow the similar distribution of the training images.

\begin{figure*}[h]
	\begin{center}
		\includegraphics[width=0.9\linewidth]{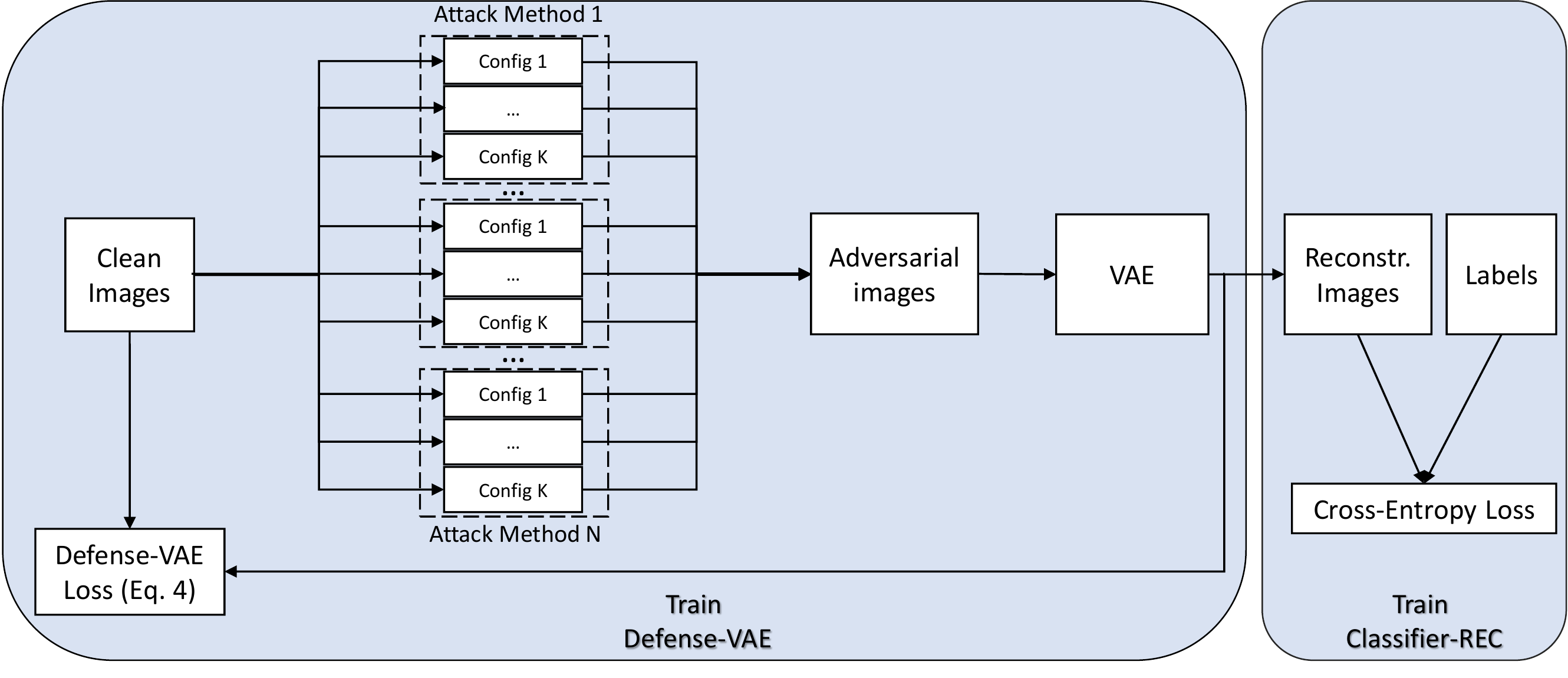}
	\end{center}\vspace{-12pt}
	\caption{Training pipeline of Defense-VAE. Defense-VAE (left) and Classifier-REC (right) can be trained separately, or jointly end-to-end (from scratch or by fine-tuning). See text for more details.}
	\label{fig:train}
\end{figure*}

\begin{figure*}[h]
	\begin{center}
		\includegraphics[width=0.9\linewidth]{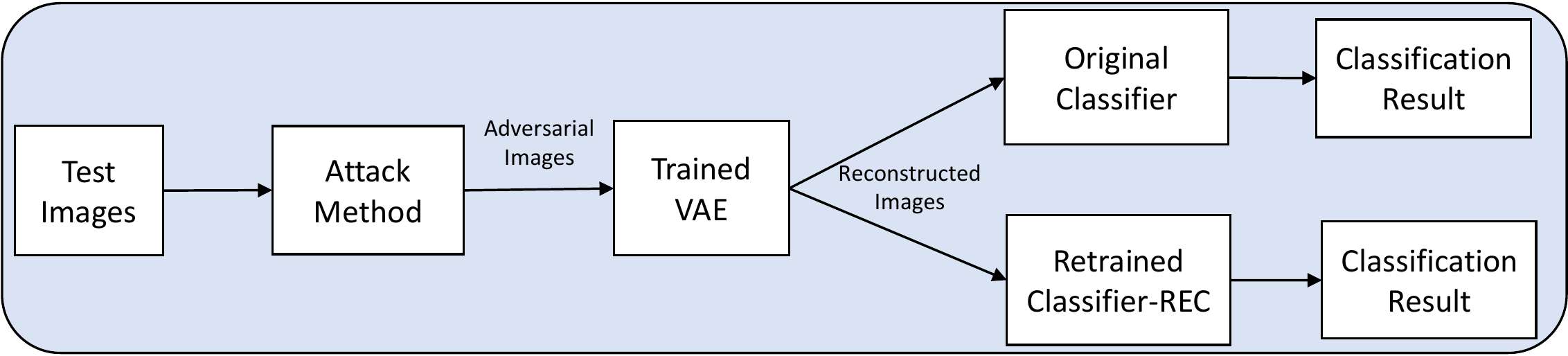}
	\end{center}\vspace{-12pt}
	\caption{Test pipeline of Defense-VAE}
	\label{fig:test}
\end{figure*}\vspace{-8pt}

\subsection{Defense-VAE}\label{sec:defense-vae}
VAE is typically trained to reproduce the same image from an input image. As for adversarial defense, reproducing the same adversarial images is an undesirable task as the adversarial perturbations may be preserved during the image reconstruction. Instead, in Defense-VAE, we modify the encoder and the decoder of the latent variable model as follows:
\begin{align}
\boldsymbol{z} \sim \operatorname { Enc } ( \boldsymbol { \hat{x} } ) = q ( \boldsymbol { z } | \boldsymbol { \hat{x} } ) , \quad \boldsymbol { x } \sim \operatorname { Dec } ( \boldsymbol { z } ) = p ( \boldsymbol { x } | \boldsymbol { z } ),
\end{align}
where $\boldsymbol{\hat{x}}=\boldsymbol{x}+\boldsymbol{\delta}$ is an adversarial image with the perturbation $\boldsymbol{\delta}$ added on top of a clean image $\boldsymbol{x}$. This adversarial image is encoded to a latent variable $\bs{z}$, which is decoded to the underlying clean image $\bs{x}$. Accordingly, the training loss of Defense-VAE is updated as follows:
\begin{align}
\mathcal { L } _ { \mathrm { Defense-VAE } } &= - \mathbb { E } _ { q ( \boldsymbol { z } | \boldsymbol {\hat{ x }} ) } \left[ \log \frac { p ( \boldsymbol { x } | \boldsymbol { z } ) p ( \boldsymbol { z } ) } { q ( \boldsymbol { z } | \boldsymbol {\hat{ x }} ) } \right]\label{eq:defense-vae}\\ \nonumber
& = - \mathbb { E } _ { q ( \boldsymbol { z } | \boldsymbol {\hat{ x }} ) } [ \log p ( \boldsymbol { x } | \boldsymbol { z } ) ]  +
D _ { \mathrm { KL } } ( q ( \boldsymbol { z } | \boldsymbol {\hat{ x }} ) \| p ( \boldsymbol { z } ) ),
\end{align}
where the input to Defense-VAE is an adversarial image $\boldsymbol{\hat{x}}=\boldsymbol{x}+\boldsymbol{\delta}$, and the expected output is the underlying clean image $\bs{x}$. The compatibility between input and output pair is measured by the loss function~\ref{eq:defense-vae}. 

To train the Defense-VAE model, we can generate adversarial images given any clean image from a training set. Since there are many different adversarial attack algorithms and for each attack algorithm we can generate multiple adversarial images with different configurations, we can in principle generate an unlimited amount of training pairs for Defense-VAE, i.e., multiple adversarial images can be mapped to one clean image. The detailed training pipeline is demonstrated in Figure~\ref{fig:train} (left). Being an effective approach of generating sufficient training pairs for Defense-VAE, using multiple attack algorithms to produce adversarial training examples will also boost the capability of Defense-VAE to counter an ensemble of adversarial attacks and make Defense-VAE a generic defense algorithm that is robust to a wide range of attacks. As we will discuss later, this ensemble training strategy entails Defense-VAE superior defense capability over Defense-GAN.

Once the Defense-VAE model is trained, we can also use the reconstructed images from Defense-VAE to retrain the downstream CNN classifiers Figure~\ref{fig:train} (right). As we will see later, the retrained CNN classifier can further boost the defense accuracy over the original CNN classifier.

We can also train the whole pipeline end to end from scratch or finetuning from pre-trained VAE and CNN classifier by optimizing the joint loss function:
\begin{align}
\mathcal { L }_{\mathrm{End-to-End}}=\mathcal { L } _ { \mathrm { Defense-VAE } } + \lambda \mathcal { L }_{\mathrm{Cross-Entropy}}.
\label{eq:finetune}
\end{align}
As we will see from the experiments, this end to end training can boost the defense accuracy even further.

After training the Defense-VAE model and potentially retraining CNN classifiers or end-to-end finetuning the whole pipeline, we can use the trained Defense-VAE to purge the adversarial perturbations from any contaminated images, and the reconstructed images are then fed to the original CNN classifier or retrained CNN classifier for the final image classification. This test pipeline is shown in Figure~ \ref{fig:test}.

\section{Related Work} \label{sec:related}
Adversarial attacks and defenses is one of the active research areas in deep learning, with tens of different attack and defense algorithms developed in the past few years. For a general introduction to this exciting research area and the related terminologies, we refer the readers to~\cite{aml, yuan2017adversarial, samangouei2018defensegan} for more details. Here we will focus on the defense algorithms that are most closely related to Defense-VAE.
 
Defending against adversarial attacks is a challenging task. Different types of defense algorithms~\cite{papernot2016distillation,moosavi2016deepfool} have been proposed in the past few years. The first type of defense algorithms~\cite{dziugaite2016study,guo2017countering,luo2015foveationbased} augments the training data to make the DNN model resilient to the trained adversarial attacks. The second type of defense algorithms~\cite{gu2014deep,ross2017improving,Lyu_2015,nguyen2017learning,nayebi2017biologically,gao2017deepcloak} modifies the training process by introducing regularization to the objective functions. The third type of defense algorithms~\cite{akhtar2017defense,Xu_2018,GuoRanCis17} attempts to remove the adversarial perturbations via input transformations before feeding the image to the classifier. According to this categorization, our Defense-VAE belongs to the input transformation based defense approach. In the following, we will therefore review the defense algorithms that are closely related to our work.

Adversarial training \cite{goodfellow6572explaining,kurakin2016adversarial} is a popular and well investigated defense approach against adversarial attacks. It attempts to use adversarial images as data augmentation to train a robust classifier. It shows that this method can improve the defense accuracy effectively and sometimes it can even improve the accuracy upon the model trained only on the original clean training set. However, this defense mechanism is more effective in white-box attacks than in black-box attacks due to the gradient masking problem. In Defense-VAE, we also use adversarial examples to improve the robustness of the defense model. However, instead of improving the targeted CNN classifiers directly, adversarial training is used to train a Defense-VAE model to purge adversarial perturbations for the downstream CNN classifiers.

Magnet proposed by Meng and Chen~\cite{Meng_2017} is another effective strategy to defend adversarial attacks. Magnet has two phases for defense: detector network and reformer network. Detector network learns the manifold of the normal clean images so that it can detect if an input image is an adversarial. If an image is detected as an adversarial, it will be forwarded to the reformer network, which will modify the adversarial image to the manifold of normal images. In Magnet, the reformer network is trained only on clean images with the goal of reconstructing the same clean input images, while Defense-VAE is trained on adversarial and clean image pairs with the goal of removing the adversarial perturbations from the contaminated images. 
 
Another closely related work is Defense-GAN that is proposed by Samangouei et. al. in ~\cite{samangouei2018defensegan}, where a Generative Adversarial Network (GAN)~\cite{gan} is used to reconstruct a clean image from an adversarial image. Defense-GAN firstly trains a GAN model purely on a training set of clean images, and as such it learns the distribution of the normal images. Then given an adversarial image, multiple iterations of back-propagations are used to identify a proper $\bs{z}$ from the clean image latent space, such that after decoded through the GAN generator, the reconstructed image is expected to be as close as possible to the adversarial image. Given the non-convex loss function of the GAN generator model, multiple random $\bs{z}$'s are used to initialize the back-propagation image search. Typically, given an adversarial image, Defense-GAN needs to perform $L$ iterations of back-propagation for each of $R$ random initializations, with the typical values of $L=200$ and $R=10$. As a comparison, to reconstruction a clean image, Defense-VAE can directly identify a proper $\bs{z}$ by forward-propagating an adversarial image through the VAE encoder network, and the $\bs{z}$ is subsequently used to reconstruct a clean image through the VAE-decoder network. No expensive iterative online optimization is needed in Defense-VAE. As we will discuss later, such reconstructed images are not only more accurate, but the whole process is much faster than Defense-GAN.

\section{Experiments}
We validate our algorithm on four popular image classification benchmarks: MNIST~\cite{mnist}, F-MNIST~\cite{xiao2017fashion}, CelebA~\cite{liu2015faceattributes} and CIFAR-10~\cite{cifar10}. MNIST and F-MNIST are two gray-level image datasets, each containing 60,000 training images and 10,000 test images with the size of $28 \times 28$. While MNIST consists of 10 hand-written digits, F-MNIST contains 10 different articles, e.g., shoes, shirts, etc. CelebA contains 202,599 RGB images of human faces, split into training and test sets. We use this dataset for binary classification to distinguish if a face image is from a male or a female. CIFAR-10 contains 10 classes of RGB images of the size of $ 32\times32$, in which 50,000 images are for training and 10,000 images are for test. 

We consider both the white-box attacks and the black-box attacks to test the defense performance of our algorithm. For the white-box attacks, FGSM~\cite{goodfellow6572explaining}, Randomized FGSM~\cite{kurakin2016adversarial}, and CW~\cite{carlini2017towards} attacks are used. For the black-box attacks, we train a substitute model to generate adversarial images to attack the targeted CNN classifiers. For a fair comparison, our experimental setups closely follow those of Defense-GAN~\footnote{\url{https://github.com/kabkabm/defensegan}}.

To demonstrate the generalization of our algorithm, we test our algorithms with the targeted CNN classifiers of different architectures: different number of convolutional or full-connected layers, different convolution parameters, and with/without dropout or batch normalization. For the black-box attacks, different architectures are also considered for the substitute models. When we present results, we denote the targeted model as A, B, C, D and the substitute model as B, E. Detailed network architectures of the VAE model, the targeted CNN classifiers and their substitutes are summarized in Appendix~\ref{app:arch}. 

For the defense algorithms, we compare our algorithm with Adversarial Training~\cite{goodfellow6572explaining, kurakin2016adversarial}, MagNet~\cite{Meng_2017} and Defense-GAN~\cite{samangouei2018defensegan}. All of our experiments are performed on NVIDIA Titan-Xp GPUs. Our source code can be found at \url{https://github.com/lxuniverse/defense-vae}.

\subsection{Results on White-box Attacks}
First, we test our algorithm on three types of white-box attacks: FGSM, RAND-FGSM and CW attacks. The targeted CNN models are trained on the original training dataset for 10 epochs until convergence. Then for each clean training image we generate 12 different adversarial images by using 3 different white-box attack algorithms, each with 4 different configurations. For FGSM and RAND-FGSM, 4 different $\epsilon= 0.25, 0.3, 0.35$ and $0.4$  are used. For the CW attack, 4 different learning rates $lr=6, 8, 10$ and $12$ are used. We combine these adversarial images and the original clean images to form the input and output pairs to train the Defense-VAE model. We initialize the weights of VAE with the normal distribution of $\mathcal{N}(0, 0.02)$ for the convolutional layers and $\mathcal{N}(1, 0.02)$ for the batch normalization layers. We note that usually 5 epochs are required for the Defense-VAE models to converge. 

Additionally, we use the reconstructed images of Defense-VAE to retrain the CNN classifiers to improve the classification accuracy. Although the original CNN classifiers have already yielded very competitive performance compared with Defense-GAN, we note that retraining CNN classifiers for Defense-VAE can further strengthen the defense accuracy notably. Interestingly, the authors of Defense-GAN reported that for Defense-GAN retraining of CNN classifiers has negligible impact to the defense accuracy, while this is not true for Defense-VAE.

As discussed in Sec.~\ref{sec:defense-vae}, we can also train the whole pipeline end to end by optimizing the joint loss function~\ref{eq:finetune} directly. This can be done through two approaches: (1) randomly initialize the VAE and CNN classifier model parameters and train the whole pipeline from scratch, and (2) pretrain VAE and CNN classifier separately and finetune the whole pipeline. Our experiments show that both approaches are almost equally effective, with the finetuning yielding slightly better results. We therefore only report the finetuning results in the following.

\begin{figure}[t]	
	\begin{center}
		\includegraphics[width=0.8\linewidth]{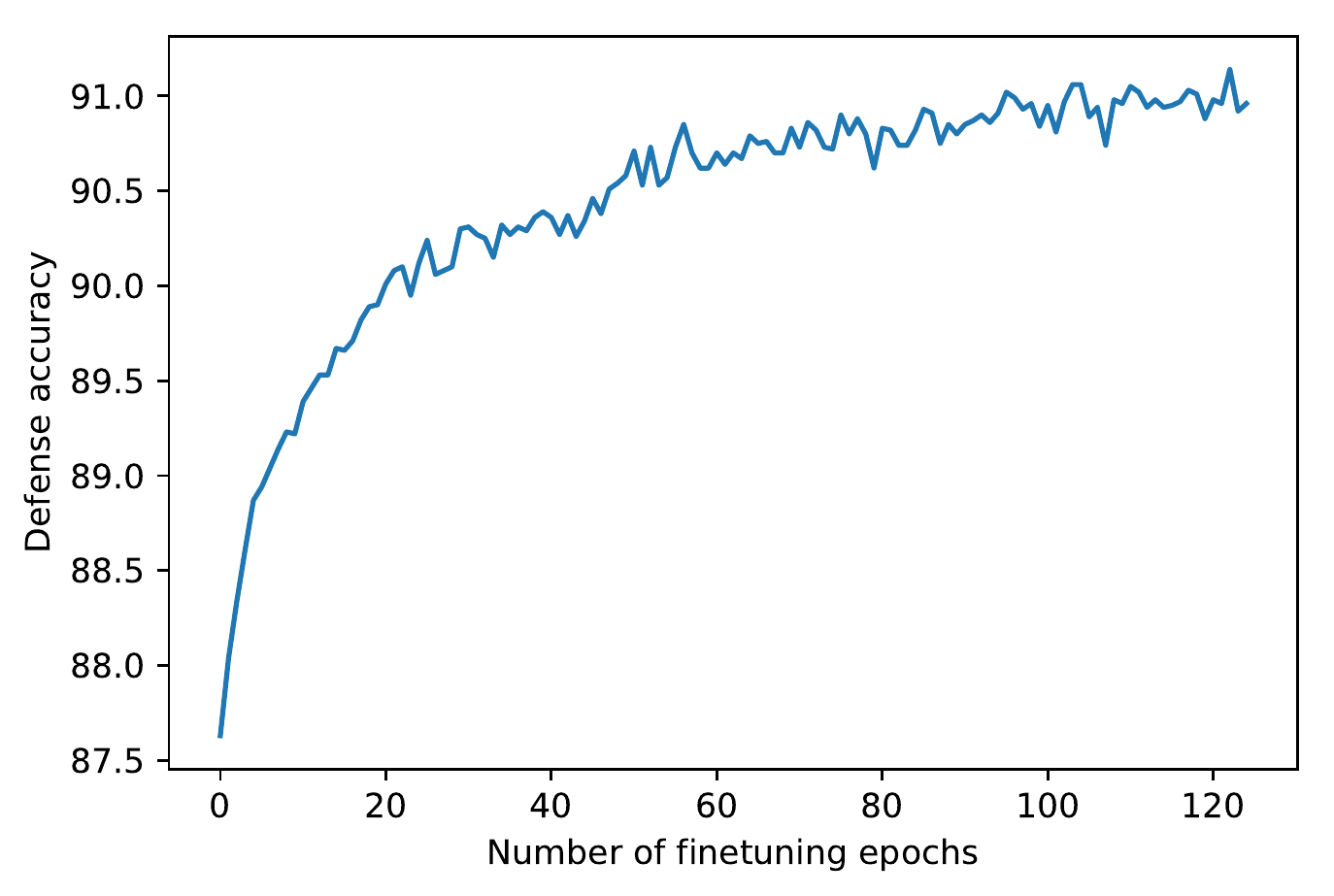}
	\end{center}\vspace{-20pt}
	\caption{The end-to-end finetuning can boost the defense accuracy even further, and yields the strongest defense model.}
	\label{fig:finetune}\vspace{-16pt}
\end{figure}

To demonstrate the effectiveness of this end-to-end finetuning approach, we provide one typical learning curve of the finetuning process in Figure~\ref{fig:finetune}, where the adversarial attacks are generated by FGSM with $\epsilon=0.3$. Starting from separately pretrained VAE model and CNN classifier (a.k.a., Defense-VAE-REC), we finetune the whole pipeline by optimizing the joint loss function~\ref{eq:finetune}. As we can see, the end-to-end finetuning boosts the defense accuracy by about 4\% over the Defense-VAE model.

\begin{table*}[h!]
	\begin{center}
		\scalebox{1.00}{
		\begin{tabular}{|l|lllllllll|}
		\hline
			Attack                                                                                    & \begin{tabular}[c]{@{}l@{}}Classifier\\ Model\end{tabular} 
			& \begin{tabular}[c]{@{}l@{}}No \\ Attack\end{tabular} 
			& \begin{tabular}[c]{@{}l@{}}No\\ Defense\end{tabular} 
			& \begin{tabular}[c]{@{}l@{}}Defense\\ VAE\end{tabular} 
			& \begin{tabular}[c]{@{}l@{}}Defense\\ VAE-REC\end{tabular} 
			& \begin{tabular}[c]{@{}l@{}}Defense\\ VAE-E2E\end{tabular} 
			& \begin{tabular}[c]{@{}l@{}}Defense\\ GAN\end{tabular} 
			& MagNet 
			& \begin{tabular}[c]{@{}l@{}}Adv. Tr.\\ $\epsilon=0.3$\end{tabular} \\ \hline			
			\multirow{4}{*}{\begin{tabular}[c]{@{}l@{}}FGSM\\ $\epsilon=0.3$\end{tabular}} 
			& A     & 90.85       & 9.18       & 86.9     &89.03  & 91.02 & 87.9  & 8.9  & 79.7      \\
			& B     & 71.62       & 15.89      & 70.88    &74.41  & 77.86 & 62.9  & 16.8 & 13.6       \\
			& C     & 90.78       & 8.68       & 85.8     &89.72  & 90.85  & 89.6  & 11.0 & 80.4        \\
			& D     & 86.94       & 8.51       & 85.36    &87.09  & 89.26  & 87.5  & 9.9  & 69.8          \\ \hline
			\multirow{4}{*}{\begin{tabular}[c]{@{}l@{}}RAND\\ FGSM\\ $\epsilon=0.3$\\ $\alpha=0.05$\end{tabular}} 
			& A     & 90.85       & 7.91       & 86.42    &88.91  & 90.57 & 88.8   & 9.6  & 44.7        \\
			& B     & 71.62       & 13.14      & 71.12    &73.91  & 77.09 & 66.1    & 16.1   & 11.9         \\
			& C     & 90.78       & 5.48       & 86.42    &89.38  & 90.28 & 89.3    & 11.2   & 69.9         \\
			& D     & 86.94       & 7.79       & 85.77    &87.18  & 88.97 & 86.2    & 10.4   & 62.6     \\ \hline
			\multirow{4}{*}{\begin{tabular}[c]{@{}l@{}}CW\\ $l_2$ norm\end{tabular}}                    
			& A     & 90.85       & 11.67      & 81.81    &86.99  & 88.54  & 89.6    & 6.0    & 15.7     \\
			& B     & 71.62       & 18.74      & 67.43    &73.69  & 74.72  & 65.6    & 13.1   & 11.8   \\
			& C     & 90.78       & 7.70        & 78.64   &87.47  & 88.69   & 89.6    & 8.4    & 10.7       \\
			& D     & 86.94       & 9.35       & 64.38    &86.21  & 87.83 & 87.5    & 6.9    & 14.9     \\ \hline
		Average &  & 84.05 & 10.34 & 79.24  & 84.50 & \textbf{86.31} & 82.55 & 10.69 & 40.48\\\hline
		\end{tabular}
		}
	\end{center}\vspace{-4pt}

	\begin{center}
		\scalebox{1.00}{
		\begin{tabular}{|l|lllllllll|}
		\hline
			Attack  & \begin{tabular}[c]{@{}l@{}}Classifier\\ Model\end{tabular} 
			& \begin{tabular}[c]{@{}l@{}}No \\ Attack\end{tabular} 
			& \begin{tabular}[c]{@{}l@{}}No\\ Defense\end{tabular} 
			& \begin{tabular}[c]{@{}l@{}}Defense\\ VAE\end{tabular} 
			& \begin{tabular}[c]{@{}l@{}}Defense\\ VAE-REC\end{tabular} 
			& \begin{tabular}[c]{@{}l@{}}Defense\\ VAE-E2E\end{tabular} 
			& \begin{tabular}[c]{@{}l@{}}Defense\\ GAN\end{tabular} 
			& MagNet & \begin{tabular}[c]{@{}l@{}}Adv. Tr.\\ $\epsilon=0.3$\end{tabular} \\ \hline
			\multirow{4}{*}{\begin{tabular}[c]{@{}l@{}}FGSM\\ $\epsilon=0.3$\end{tabular}} 
			& A     &99.15    & 14.65  & 98.29    & 98.98  & 99.28 & 98.8   & 19.1   &  65.1      \\
			& B     &96.10     & 1.81   & 95.92    & 95.97 & 96.91 & 95.6   & 8.2    &  6.0     \\
			& C     &99.08    & 29.53  & 98.41    & 98.91  & 99.24 & 98.9   & 16.3   &  78.6      \\
			& D     &97.87    & 4.33   & 97.56    & 98.16  & 98.05 & 98.0   & 9.4    &  73.2     \\ \hline
			\multirow{4}{*}{\begin{tabular}[c]{@{}l@{}}RAND\\ FGSM\\ $\epsilon=0.3$\\ $\alpha=0.05$\end{tabular}} 
			& A     &99.15    & 8.65   & 98.40    & 99.08  & 99.34  & 98.8   & 17.1   & 77.4        \\
			& B     &96.10     & 1.65   & 95.83   & 96.04  & 96.87  & 94.4   & 9.1    & 13.8      \\
			& C     &99.08    & 5.99   & 98.33    & 98.87  & 99.35  & 98.5   & 15.1   & 90.7        \\
			& D     &97.87    & 3.25   & 97.81    & 98.3   & 98.05  & 98.0   & 11.5    & 53.9        \\ \hline
			\multirow{4}{*}{\begin{tabular}[c]{@{}l@{}}CW\\ $l_2$ norm\end{tabular}}                    
			& A     &99.15    & 8.45   & 92.69    & 95.12  & 96.95  & 98.9   & 3.8    &  7.7       \\
			& B     &96.10     & 3.00    & 87.66  & 88.56  & 95.08   & 91.6   & 3.4    &  28.0     \\
			& C     &99.08    & 5.53   & 94.46    & 96.05  & 96.44  & 98.9   & 2.5    &  3.1       \\
			& D     &97.87    & 3.92   & 83.42    & 89.46  & 95.71  & 98.3   & 2.1    &  1.0   \\ \hline
		Average &  & 98.05 & 7.56 & 94.90  & 96.13 &  \textbf{97.61}  &97.39 & 9.80 & 27.38\\\hline
		\end{tabular}
		}
	\end{center}
	\caption{Classification accuracies of different defense methods under FGSM, RAND-FGSM and CW white-box attacks on the (top) F-MNIST and (bottom) MNIST image classification benchmarks. The defense accuracies of Defense-GAN, MagNet, and Adversarial Training are from Defense-GAN~\cite{samangouei2018defensegan}. Results on CelebA and CIFAR-10 have the same pattern as above. Details can be found in Appendix~\ref{app:exp}.}
	\label{tab:white}\vspace{-12pt}
\end{table*}

Table~\ref{tab:white}  reports the defense accuracies of Defense-VAE on three different white-box attacks: FGSM, RAND-FGSM and CW attacks. As a comparison, we also include the results of Defense-GAN, MagNet and Adversarial Training under the same experimental setups; for those results, we import them directly from the Defense-GAN paper~\cite{samangouei2018defensegan}. As we can see, Defense-VAE and Defense-GAN are very competitive to each other,  and outperform all the other defense algorithms by significant margins on all four benchmarks. Defense-VAE achieves superior performance over Defense-GAN, and can recover almost all the accuracy losses due to the adversarial attacks. We also note that retraining CNN classifiers (Defense-VAE-REC) and finetuning (Defense-VAE-E2E) can further improve the defense accuracies beyond the original CNN classifiers (Defense-VAE) by a notable margin, with the finetuning yielding the strongest defense against adversarial attacks.

\vspace{-8pt}
\subsection{Robustness under Untrained Attacks}
In principle we can train Defense-VAE on all known adversarial attacks to best counter possible attacks in test. However, in reality new attacks are constantly invented; it's almost certain that after the deployment of Defense-VAE, some new adversarial attacks will emerge and Defense-VAE has never been trained on those attacks. To investigate the robustness of Defense-VAE in this circumstance, in this part of the experiments we train Defense-VAE on two attacks and test its defense capability against the third untrained attack. Again, three adversarial attacks are considered: FGSM, RAND-FGSM and CW, which gives three possible combinations that are shown in Table~\ref{tab:2_1}. As we can see, Defense-VAE is very robust for the first two attacks: FGSM and RAND-FGSM as the defense accuracies largely remain the same as it's trained on all three attacks. But for the CW attack, Defense-VAE is less robust, manifested by the significant accuracy loss compared to the Defense-VAE trained on all three attacks. Indeed, the CW attack is considered a much stronger attack and could have a very distinct attack pattern to that of FGSM and RAND-FGSM.
We therefore incorporate Deepfool~\cite{moosavi2016deepfool} to the training of Defense-VAE to counter the untrained CW attack since DeepFool and CW have very similar attack patterns. The results in parentheses show that this is indeed the case and Defense-VAE again can recover the most accuracy losses under untrained CW attack. 

\begin{table}[!h]\vspace{-8pt}
	\begin{center}
		\scalebox{1.00}{
		\begin{tabular}{|l|c|c|c|}
		\hline
			Attack & Classifier  & Trained on other 2 & Trained on 3   \\ \hline
			\multirow{4}{*}{FGSM}
			& A &87.34   &89.03  \\
			& B &73.38   &74.41  \\
			& C &88.03   &89.72  \\
			& D &86.49   &87.09  \\ \hline
			\multirow{4}{*}{\begin{tabular}[c]{@{}l@{}}RAND\\ FGSM\end{tabular}} 
			& A &87.30    &88.91  \\
			& B &73.59   &73.91  \\
			& C &88.19   &89.38  \\
			& D &86.73   &87.18  \\ \hline
			\multirow{4}{*}{CW}                                                  
			& A & 43.48 (85.06)  &86.99  \\
			& B & 34.52 (71.64)  &73.69  \\
			& C & 44.45 (85.22)  &87.47  \\
			& D & 30.77 (84.69)  &86.21   \\\hline
		\end{tabular}
		}
	\end{center}
	\caption{Defense accuracy of Defense-VAE when it's trained on two attacks but is used to defend another attack. The results in parentheses are the accuracies after incorporating DeepFool~\cite{deepfool} as additional adversarial training examples for Defense-VAE.}
	\label{tab:2_1}\vspace{-28pt}
\end{table}

%\begin{table}[!h]
%	\begin{center}		
%		\begin{tabular}{c|c|c}
%			$\epsilon$    & MNIST& F-MNIST \\ \hline
%			0.10 &  98.95    &86.04                 \\
%			0.15 &  98.67    & 86.32         \\
%			0.20 &  98.58    & 86.39              \\
%			0.25 &  98.44    & 86.51                  \\
%			0.30 &  98.29    & 86.36                                                        
%		\end{tabular}
%	\end{center}\vspace{-4pt}
%	\caption{Defense accuracy of Defense-VAE with Model A under the FGSM attack with various noise level $\epsilon$ when VAE is trained only on $\epsilon=0.3$.}
%	\label{tab:deff_eps}\vspace{-8pt}
%\end{table}

%Another interesting defense scenario is: what if Defense-VAE were tested on the same type of attacks but with different attack configurations? To investigate this, we train Defense-VAE on the FGSM attack with $\epsilon=0.3$ and test its defense accuracies with different $\epsilon$.  We validate this on MNIST and Fashion-MNIST, with the results shown in Table~\ref{tab:deff_eps}. It shows that that Defense-VAE is very robust to the untrained FGSM configurations as the defense accuracies largely remain the same under the trained attacks, e.g., with $\epsilon=0.3$. 

\vspace{-8pt}
\subsection{Results on Black-box Attacks}
Next, we test the defense capability of Defense-VAE under black-box attacks on the MNIST and F-MNIST datasets. We train the targeted CNN model on the training set for 10 epochs with the batch size of 100 and the learning rate of $10^{-3}$ until convergence. Then the substitute model is trained with 150 images from the test set with the labels predicted by the targeted CNN classifier. 

\begin{table*}[!h]
	\begin{center}
		\scalebox{1.0}{
		\begin{tabular}{|l|l|l|l|l|l|l|l|l|}
		\hline
			\begin{tabular}[c]{@{}l@{}}Classifier/\\ Substitute\end{tabular} 
			& \begin{tabular}[c]{@{}l@{}}No\\ Attack\end{tabular} 
			& \begin{tabular}[c]{@{}l@{}}No\\ Defense\end{tabular} 
			& \begin{tabular}[c]{@{}l@{}}Defense-\\ VAE\end{tabular} 
			& \begin{tabular}[c]{@{}l@{}}Defense-\\ VAE-REC\end{tabular} 
			& \begin{tabular}[c]{@{}l@{}}Defense-\\ VAE-E2E\end{tabular} 
			& \begin{tabular}[c]{@{}l@{}}Defense-\\ GAN\end{tabular} 
			& \begin{tabular}[c]{@{}l@{}}MagNet\end{tabular} 
			&  \begin{tabular}[c]{@{}l@{}}Adv. Tr.\\ $\epsilon$ = 0.3\end{tabular} \\ \hline
			A/B     & 90.85 & 37.92 & 83.69 & 86.64 & 86.39 & 58.60 & 54.04 & 73.93   \\ 
			A/E     & 90.85 & 24.94 & 76.97 & 83.02 & 83.61 & 47.90 & 33.11 & 69.45   \\ 
			B/B     & 71.62 & 17.61 & 73.66 & 72.42 & 75.22 & 49.40 & 38.12 & 31.77   \\ 
			B/E     & 71.62 & 13.44 & 69.29 & 69.36 & 71.78 & 37.20 & 31.19 & 26.17   \\ 
			C/B     & 90.78 & 39.14 & 83.64 & 86.88 & 87.67 & 52.89 & 46.64 & 77.91   \\ 
			C/E     & 90.78 & 22.89 & 76.27 & 80.16 & 80.32 & 48.71 & 30.16 & 75.04   \\ 
			D/B     & 86.94 & 32.87 & 80.31 & 85.80 & 84.78 & 57.79 & 54.78 & 61.72   \\ 
			D/E     & 86.94 & 23.51 & 70.66 & 79.48 & 77.53 & 40.07 & 33.96 & 50.93   \\ \hline
			Average & 85.05 & 26.54 & 76.81 & 80.47 &\textbf{80.91} & 49.07 & 40.25 & 58.37 \\\hline
		\end{tabular}
		}
	\end{center}
%\end{table*}
%\begin{table*}[!htbp]

	\begin{center}
		\scalebox{1.00}{
		\begin{tabular}{|l|l|l|l|l|l|l|l|l|}
		\hline
			\begin{tabular}[c]{@{}l@{}}Classifier/\\ Substitute\end{tabular} 
			& \begin{tabular}[c]{@{}l@{}}No\\ Attack\end{tabular} 
			& \begin{tabular}[c]{@{}l@{}}No\\ Defense\end{tabular} 
			& \begin{tabular}[c]{@{}l@{}}Defense-\\ VAE\end{tabular} 
			& \begin{tabular}[c]{@{}l@{}}Defense-\\ VAE-REC\end{tabular} 
			& \begin{tabular}[c]{@{}l@{}}Defense-\\ VAE-E2E\end{tabular} 
			& \begin{tabular}[c]{@{}l@{}}Defense-\\ GAN\end{tabular} 
			&  \begin{tabular}[c]{@{}l@{}}MagNet\end{tabular} 
			&  \begin{tabular}[c]{@{}l@{}}Adv. Tr.\\ $\epsilon$ = 0.3\end{tabular}\\ \hline
			A/B  & 99.15    & 65.89   &  98.68  & 98.71 & 99.16 & 93.12   & 69.37   & 96.54 \\ 
			A/E  & 99.15    & 76.32   &  98.64  & 98.92 & 99.19 & 91.39   & 67.10   & 96.68 \\ 
			B/B  & 96.10    & 14.40   &  95.89  & 95.95 & 96.71 & 90.57   & 56.87   & 20.92 \\ 
			B/E  & 96.10    & 26.48   &  96.26  & 95.81 & 97.09 & 88.41   & 46.27   & 11.20 \\ 
			C/B  & 99.08    & 60.74   &  97.91  & 98.02 & 99.15 & 93.57   & 75.71   & 98.34 \\ 
			C/E  & 99.08    & 72.73   &  98.30  & 98.59 & 99.28 & 92.23   & 67.60   & 98.43 \\ 
			D/B  & 97.87    & 33.36   &  97.68  & 98.22 & 97.85 & 92.72   & 68.17   & 76.67 \\ 
			D/E  & 97.87    & 39.95   &  97.72  & 98.22 & 97.69 & 91.64   & 60.73   & 76.76 \\ \hline
			Average & 98.05 & 48.73   & 97.63 & 97.81  & \textbf{98.27} & 91.71 & 63.98 & 71.92\\\hline
		\end{tabular}
		}
	\end{center}\vspace{-4pt}
\caption{Classification accuracies of different defense methods under FGSM black-box attacks on different image classification benchmarks: (top) F-MNIST, (bottom) MNIST. The defense accuracies of Defense-GAN, MagNet, and Adversarial Training are from the Defense-GAN paper~\cite{samangouei2018defensegan}. Results on CIFAR-10 have the same pattern as above. Details can be found in Appendix~\ref{app:exp}.}
	\label{tab:black}\vspace{-20pt}
\end{table*}

In the black-box attacks, Defense-VAE, as a defender, has no prior knowledge of the trained substitute model. Thus, we can only train Defense-VAE on the white-box attacks. Therefore, the same Defense-VAE model trained from the experiments of white-box attacks is used to defend the black-box attacks.~\footnote{In other words, we just need to train one Defense-VAE to defend both white-box and black-box attacks.} In this experiment, 4 targeted CNN classifiers: A, B, C, and D, and 2 substitute models: B and E are considered, and this produces 8 possible Classifier/Substitute combinations. In this part of experiments, only the black-box FGSM attack is considered, with the results on MNIST and F-MNIST reported in Table~\ref{tab:black}. As a comparison, we also include the results of Defense-GAN, MagNet and Adversarial Training under the same experimental setups; again, for this set of results, we import them directly from the Defense-GAN paper~\cite{samangouei2018defensegan}. As we can see, on both datasets Defense-VAE outperforms Defense-GAN and all other defense algorithms by significant margins. In particular, on F-MNIST, Defense-VAE improves the accuracy over Defense-GAN by about 30\%. Also, as in the white-box attack experiments, retrained CNN classifiers (Defense-VAE-REC) and finetuning (Defense-VAE-E2E) can further boost the defense accuracies over the original CNN classifiers (Defense-VAE) by a notable margin, with the end-to-end finetuning yielding the best defense accuracies among all the methods.

\subsection{Why is Defense-VAE so effective?}
\begin{figure}[h!]\vspace{-8pt}
	\begin{center}
		\includegraphics[width=0.85\linewidth]{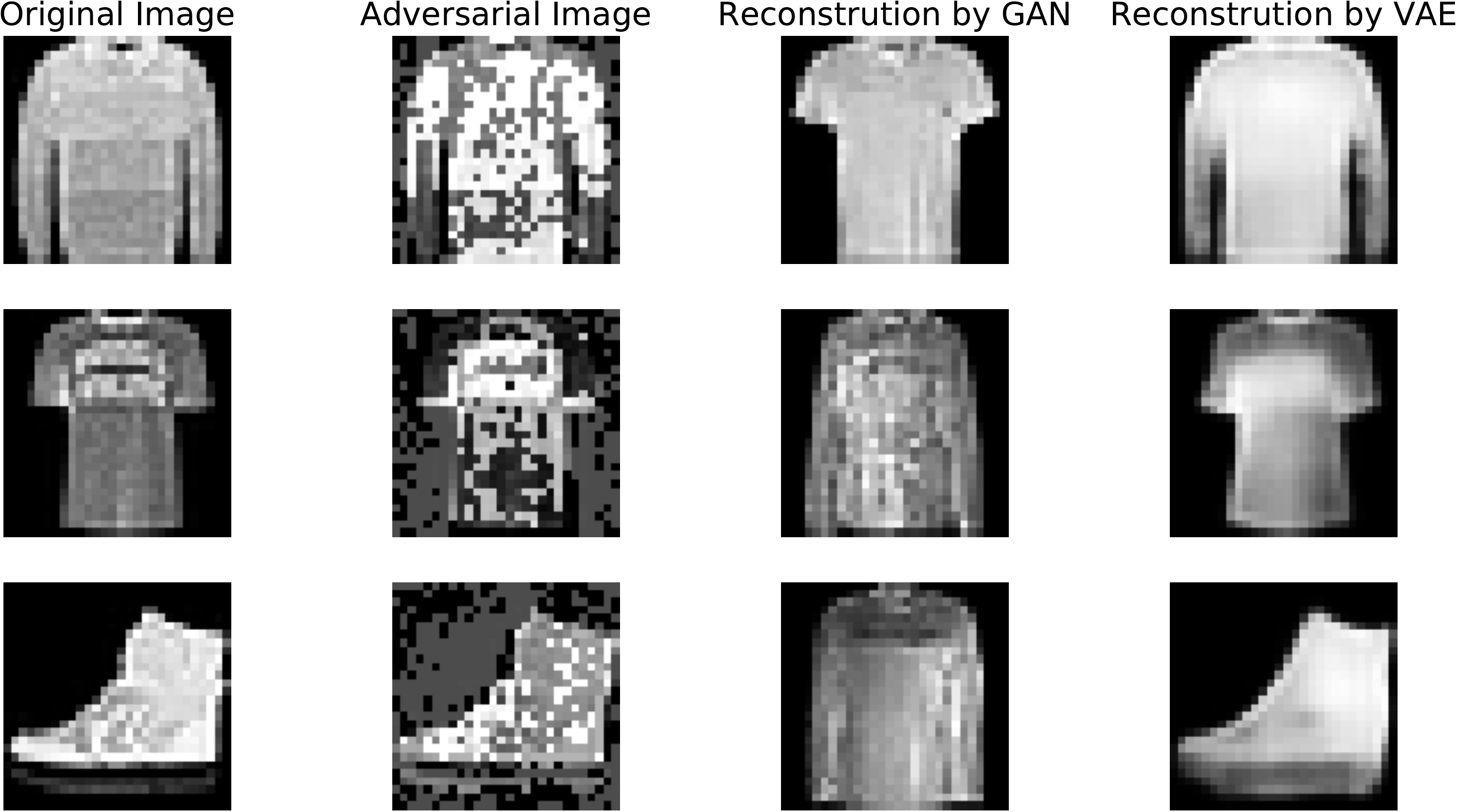}
	\end{center}\vspace{-12pt}
	\caption{The example reconstructions by Defense-VAE and Defense-GAN from the black-box FGSM attacks on F-MNIST: (a) original images; (b) adversarial images; (c) reconstruction by Defense-GAN; (d) reconstruction by Defense-VAE.}\vspace{-8pt}
	\label{fig:rec}\vspace{-16pt}
\end{figure}

The results above demonstrated superior performance of Defense-VAE over Defense-GAN. For the black-box FGSM attack, the former even outperforms the latter by about 30\%. To understand why Defense-VAE can have such a large leap, we investigate the reconstructed images by Defense-VAE and Defense-GAN in this experimental setup, i.e., the black-box FGSM attack on F-MNIST. Figure~\ref{fig:rec} shows some typical examples from this experiment. As can be seen, the reconstructed images from Defense-VAE often preserve the correct class information of their underlying clean images, while Defense-GAN has a harder time to identify a correct reconstruction even though it searches for the right $\bs{z}$ from $R$ random initializations and optimizes in $L$ back-propagations, with typical $R=10$ and $L=200$. As we discussed in Sec.~\ref{sec:related}, Defense-VAE identifies a proper $z$ directly by forward-propagating the input adversarial image through the VAE-encoder, and reconstructs a high quality denoised image through the VAE-decoder, and no online iterative optimization is involved.

\subsection{Defense Speed}
Besides the superior defense accuracy of Defense-VAE, another advantage of Defense-VAE is its superior defense speed over Defense-GAN. As discussed above, to identify a high quality reconstruction, Defense-VAE doesn't need expensive online iterative optimizations, while Defense-GAN requires $L$ iterative back-propagations with $R$ random restarts. To have a quantitative speed comparison between Defense-VAE and Defense-GAN, we calculate their reconstruction times on 1000 adversarial images from F-MNIST, with the results reported in Table~\ref{tab:running_time}, where different $R$ and $L$ configurations are considered.

As we can see, compared to the default Defense-GAN configuration, i.e., $L=200$ and $R=10$, Defense-VAE is about 50x faster than Defense-GAN. Moreover, as $L$ and $R$ increase, Defense-GAN generally has a slightly better defense accuracy, but the run time also increases linearly as $\mathcal{O}(L\times R)$. The constant run-time complexity of Defense-VAE makes it widely deployable in real-time security-sensitive systems.
\vspace{-4pt}

\begin{table}[t]
	\begin{center}
		\begin{tabular}{|l|l|c|}
		\hline
			\multicolumn{2}{|c|}{Defense Method}     & \begin{tabular}[c]{@{}l@{}}Run Time on\\ 1000 Images (s)\end{tabular} \\ \hline
			\multicolumn{2}{|l|}{Defense-VAE}               & 9.03                                                                   \\ \hline
			\multirow{4}{*}{Defense-GAN} & L$^*$ = 200, R$^*$ = 10 & 441.81                                                                 \\ \cline{2-3} 
			& L = 400, R = 10 & 875.48                                                                 \\ \cline{2-3} 
			& L = 200, R = 20 & 876.10                                                                 \\ \cline{2-3} 
			& L = 400, R = 20 & 1720.13                                                               \\ \hline
		\end{tabular}
	\end{center}\vspace{-4pt}
	\caption{Run Time Comparison between Defense-VAE and Defense-GAN, where $^*$ denotes Defense-GAN recommended configuration.}
	\label{tab:running_time}\vspace{-20pt}
\end{table}

\vspace{-4pt}
\section{Conclusion}\vspace{-4pt}
In this paper, we propose Defense-VAE, a fast and accurate defense algorithm against adversarial attacks. The algorithm is generic and can defense both white-box and black-box attacks without the need of retraining the original CNN classifier, and can further boost the defense strength by retraining or end-to-end finetuning. Compared with the state-of-the-art algorithms, in particular, Defense-GAN, our algorithm outperforms them in almost all white-box and black-box defense benchmarks. In addition, Defense-VAE is very efficient as compared to the optimization-based defense alternatives, such as Defense-GAN, as no expensive iterative online optimizations is needed. Speed test shows that Defense-VAE is about 50x faster than Defense-VAE. Given the superior defense accuracy and speed, we believe Defense-VAE is widely deployable in real-time security-sensitive systems.

\bibliographystyle{splncs04}
\bibliography{egbib}

\appendix

\section{Network Architectures}\label{app:arch} \vspace{-4pt}
The details of network architectures used in our experiments are described. Table~\ref{tab:classifier} shows the architectures of the CNN classifiers and their substitute models, which are identical to those used in Defense-GAN~\cite{samangouei2018defensegan} for a fair comparison. 
%The meanings of the notations used in the tables are described below:
%\begin{itemize}
%    \item[-] Conv($C_1, C_2, K, S, P$) refers to a convolutional layer with input channel $C_1$, output channel $C_2$, filter size $K$, stride $S$ and padding $P$. If $C_1$ is *, it equals to 1 for gray images and 3 for RGB images.%\vspace{-4pt}
%    \item[-] ConvT($C_1, C_2, K, S, P$) refers to a transposed convolutional layer with input channel $C_1$, output channel $C_2$, filter size $K$, stride $S$ and padding $P$.%\vspace{-4pt}
%    \item[-] FC($M, N$) refers to a fully-connected layer with $M$ inputs and $N$ outputs.%\vspace{-4pt}
%    \item[-] Dropout($P$) refers to a dropout layer with dropout probability $P$.%\vspace{-4pt}
%    \item[-] ReLU refers to the Rectified Linear Unit activation.%\vspace{-4pt}
%    \item[-] BN refers to a Batch Normalization layer.
%\end{itemize}

\begin{table*}[h]\vspace{-16pt}
    \begin{center}
        \begin{tabular}{|l|l|l|l|l|}
        \hline
            A                      & B                        & C                        & D                   & E                \\ \hline
            Conv(*, 64, 5, 1, 2)   & Dropout(0.2)             & Conv(*, 128, 3, 1, 1)    & FC(200)             & FC(200)          \\
            ReLU                   & Conv(*, 64, 8, 2, 5)     & ReLU                     & ReLU                & ReLU             \\
            Conv(64, 64, 5, 2, 0)  & ReLU                     & Conv(128, 64, 5, 2, 0)   & Dropout(0.5)        & FC(200)          \\
            ReLU                   & Conv(64, 128, 6, 2, 0)   & ReLU                     & FC(200)             & ReLU             \\
            Dropout(0.25)          & ReLU                     & Dropout(0.25)            & ReLU                & FC(10) + Softmax \\
            FC(128)                & Conv(128, 128, 5, 1, 0)  & FC(128)                  & Dropout(0.25)       &                  \\
            ReLU                   & ReLU                     & ReLU                     & FC(10) + Softmax    &                  \\
            Dropout(0.5)           & Dropout(0.5)             & Dropout(0.5)             &                     &                  \\
            FC(10) + Softmax       & FC(10) + Softmax         & FC(10) + Softmax         &                     &             
        \\\hline    
        \end{tabular}
    \end{center}
\caption{The architectures of the classifiers and the substitute models used in the white-box and black-box attacks.}
\label{tab:classifier}\vspace{-16pt}
\end{table*}

\begin{table*}[h]\vspace{-20pt}
    \begin{center}
        \begin{tabular}{|l|l|}
        \hline
            \textbf{Encoder}             & \textbf{Decoder}          \\ \hline
            Conv(*, 64, 5, 1, 2) + BN + ReLU      & FC(128, 4096) + ReLU               \\
            Conv(64, 64, 4, 2, 3) + BN + ReLU        & ConvT(256, 128, 4, 2, 1) + BN + ReLU  \\
            Conv(64, 128, 4, 2, 1) + BN + ReLU       & ConvT(128, 64, 4, 2, 1) + BN + ReLU    \\
            Conv(128, 256, 4, 2, 1) + BN + ReLU      & ConvT(64, 64, 4, 2, 3) + BN + ReLU    \\
            FC1(4096, 128), FC2(4096, 128) & ConvT(64, 64, 5, 1, 2) + BN + ReLU                      \\\hline
        \end{tabular}
    \end{center}
%\vspace{-4pt}
\caption{The encoder and decoder of Defense-VAE used in the experiments.}
\label{tab:vaemodel}\vspace{-24pt}
\end{table*}
Table~\ref{tab:vaemodel} shows the architecture of the Defense-VAE model used in the experiments on MNIST and F-MNIST. The architectures used for CelebA~\cite{liu2015faceattributes} and CIFAR-10~\cite{cifar10} are largely the same except that they are 1 or 2 layers deeper. 

\vspace{-4pt}
\section{Experiments on CelebA and CIFAR-10}\label{app:exp}\vspace{-4pt}
We perform the white-box and black-box attacks on CelebA~\cite{liu2015faceattributes} and CIFAR-10~\cite{cifar10} datasets, with the results provided in Tables~\ref{tab:white_celeba_cifar10} and~\ref{tab:cifar10b}. Since Defense-GAN didn't provide results on CIFAR-10, we run their code on it and make sure the experimental settings for both algorithms are the same. We didn't provide the results related to the classifier model B due to its improper configuration for CIFAR-10, e.g., model B has much more parameters due to the large convolutional kernel size (e.g., $8\times 8$) and 3 input channels.

%Similar to the results on F-MNIST~\cite{xiao2017fashion} provided in the main text, Defense-VAE outperforms Defense-GAN consistently, and retraining CNN classifiers on the reconstructions of Defense-VAE boosts the accuracy significantly. We also notice that the defense accuracies on CIFAR-10 are not as good as on CelebA. This is because CIFAR-10 is a much more challenging task than CelebA: the former is a 10-way classification task, while the latter is only a binary classification on human faces. 

\begin{table*}[h]\vspace{-8pt}
	\begin{center}
		\begin{tabular}{|l|lllllllll|}
		\hline
			Attack                                                                                    
			& \begin{tabular}[c]{@{}l@{}}Classifier\\ Model\end{tabular} 
			& \begin{tabular}[c]{@{}l@{}}No \\ Attack\end{tabular} 
			& \begin{tabular}[c]{@{}l@{}}No\\ Defense\end{tabular} 
			& \begin{tabular}[c]{@{}l@{}}Defense\\ VAE\end{tabular} 
			& \begin{tabular}[c]{@{}l@{}}Defense\\ VAE-REC\end{tabular} 
			& \begin{tabular}[c]{@{}l@{}}Defense\\ VAE-E2E\end{tabular} 
			& \begin{tabular}[c]{@{}l@{}}Defense\\ GAN\end{tabular} 
			& MagNet 
			& \begin{tabular}[c]{@{}l@{}}Adv. Tr.\\ $\epsilon=0.3$\end{tabular} \\ \hline
			\multirow{4}{*}{\begin{tabular}[c]{@{}l@{}}FGSM\\ $\epsilon=0.3$\end{tabular}} 
				& A     & 96.55   & 3.94   & 92.40    & 94.89  & 95.10   &  92.55  & 9.85   &  12.25      \\
				& B     & 93.69   & 5.20  & 90.05    & 92.45   & 92.85   &  91.40  & 9.20   &  23.45      \\
				& C     & 95.62   & 4.45   & 92.47    & 94.46  & 95.25   &  92.55  & 10.85  &  11.30     \\
				& D     & 94.89   & 5.92  & 90.05    & 93.66   & 93.91   &  92.05  & 9.75   &  77.55       \\ \hline
				\multirow{4}{*}{\begin{tabular}[c]{@{}l@{}}RAND\\ FGSM\\ $\epsilon=0.3$\\ $\alpha=0.05$\end{tabular}} 
				& A     & 96.55   & 4.04    & 92.11    & 94.56 & 95.34   & 92.80   & 11.05   & 7.00       \\
				& B     & 93.69   & 4.76   & 90.55    & 92.57  & 93.07   & 90.30   & 10.15   & 45.15      \\
				& C     & 95.62   & 5.12    & 91.70    & 93.76 & 94.15  & 92.00   & 10.45   & 10.55        \\
				& D     & 94.89   & 6.15   & 91.42    & 93.53  & 93.87   & 91.65   & 11.05   & 6.96       \\ \hline
				\multirow{4}{*}{\begin{tabular}[c]{@{}l@{}}CW\\ $l_2$ norm\end{tabular}}                    
				& A     & 96.55   & 4.94    & 93.70    & 95.07 & 95.90   & 82.10   & 9.85   & 56.90        \\
				& B     & 93.69   & 4.90   & 90.65     & 92.40 & 93.55   & 74.65   & 9.55   & 7.25      \\
				& C     & 95.62   & 8.00    & 93.28    & 94.57 & 95.92   & 79.85   & 9.85   & 26.35        \\
				& D     & 94.89   & 6.47   & 91.15    & 93.12  & 93.39   & 77.40   & 10.40  & 50.10    \\ \hline
				Average &  & 95.19 & 5.32    & 91.63   &93.75& \textbf{94.36}  & 87.44  & 10.17   & 27.90\\\hline
		\end{tabular}
	\end{center}\vspace{-4pt}
	
	    \begin{center}
        \begin{tabular}{|l|lllllll|}
        \hline
			Attack                                                                                    
		& \begin{tabular}[c]{@{}l@{}}Classifier\\ Model\end{tabular} 
		& \begin{tabular}[c]{@{}l@{}}No \\ Attack\end{tabular} 
		& \begin{tabular}[c]{@{}l@{}}No\\ Defense\end{tabular} 
		& \begin{tabular}[c]{@{}l@{}}Defense\\ VAE\end{tabular} 
		& \begin{tabular}[c]{@{}l@{}}Defense\\ VAE-REC\end{tabular} 
		& \begin{tabular}[c]{@{}l@{}}Defense\\ VAE-E2E\end{tabular} 
		& \begin{tabular}[c]{@{}l@{}}Defense\\ GAN\end{tabular}  \\ \hline
            \multirow{4}{*}{\begin{tabular}[c]{@{}l@{}}FGSM\\ $\epsilon=0.3$\end{tabular}} 
            & A     &86.52    & 2.44  & 44.86    & 48.52  & 50.72 & 51.92         \\
            & C     &87.62    & 5.05  & 43.92    & 47.29  & 47.39 & 47.84         \\
            & D     &61.76    & 8.24  & 47.75    & 50.69  & 53.36 & 33.80       \\ \hline
            \multirow{4}{*}{\begin{tabular}[c]{@{}l@{}}RAND\\ FGSM\\ $\epsilon=0.3$\\ $\alpha=0.05$\end{tabular}} 
            & A     &86.52    & 3.71  & 39.84    & 47.80  & 50.51 &  50.36       \\
            & C     &87.62    & 3.87  & 41.28    & 46.16  & 47.91 &  48.52        \\
            & D     &61.76    & 7.94  & 47.88    & 50.67  & 51.18 &  26.78         \\ \hline
            \multirow{4}{*}{\begin{tabular}[c]{@{}l@{}}CW\\ $l_2$ norm\end{tabular}}                    
            & A     &86.52    &2.34   & 38.41   & 45.91   & 49.44 & 45.62           \\
            & C     &87.62    &7.13   & 41.21   & 46.26   & 46.19 & 43.87           \\
            & D     &61.76    &7.78   & 53.32   & 55.81   & 57.21 & 20.35       \\ \hline
        Average &   &78.63    &5.39   & 44.27 & 48.79 & \textbf{50.43}      & 41.01\\\hline
        \end{tabular}
    \end{center}\vspace{-0pt}
	\caption{Classification accuracies of different defense methods under FGSM, RAND-FGSM and CW white-box attacks on CelebA and CIFAR-10.  Since the Defense-GAN paper didn’t provide the white-box attack results on CIFAR-10, we run their code and provide the results in the table.}\label{tab:white_celeba_cifar10}\vspace{-4pt}
\end{table*}

%
%\caption{Classification accuracies of different defense methods under FGSM, RAND-FGSM and CW white-box attacks on CIFAR-10. Since the Defense-GAN paper didn't provide the white-box attack results on CIFAR-10, we run their original code and provide the results in the table.}
%\label{tab:cifar10w}

\begin{table*}[h]\vspace{-0pt}
	\begin{center}
		\begin{tabular}{|l|l|l|l|l|l|l|l|l}
		\hline
			\begin{tabular}[c]{@{}l@{}}Classifier/\\ Substitute\end{tabular} 
			& \begin{tabular}[c]{@{}l@{}}No\\ Attack\end{tabular} 
			& \begin{tabular}[c]{@{}l@{}}No\\ Defense\end{tabular} 
			& \begin{tabular}[c]{@{}l@{}}Defense-\\ VAE\end{tabular} 
			& \begin{tabular}[c]{@{}l@{}}Defense-\\ VAE-REC\end{tabular} 
			& \begin{tabular}[c]{@{}l@{}}Defense-\\ VAE-E2E\end{tabular} 
			& \begin{tabular}[c]{@{}l@{}}Defense-\\ GAN\end{tabular} \\ \hline
			C/E  & 87.62    & 14.13  &  37.22   &  42.68         & 45.72 &20.24 \\ 
			D/E  & 61.76    & 10.39  &  32.60   &  38.10         & 37.18 &11.68 \\ \hline
			Average &74.69  & 12.16   & 34.91   & 40.39  & \textbf{41.45} &16.32\\\hline
		\end{tabular}
	\end{center}\vspace{-0pt}
	\caption{Classification accuracies under FGSM black-box attacks on CIFAR-10.  Since the Defense-GAN paper didn’t provide the black-box attack results on CIFAR-10, we run their code and provide the results in the table.}
	\label{tab:cifar10b}
\end{table*}

\end{document}